%% file: root.tex
\documentclass[letterpaper, 10 pt, conference]{ieeeconf}
\IEEEoverridecommandlockouts
\IEEEoverridecommandlockouts                              

\usepackage{graphics} 
\usepackage{epsfig} 
\usepackage{mathptmx} 
\usepackage{times} 
\usepackage{amsmath} 
\usepackage{amssymb}  
\usepackage{graphicx}
\usepackage{cite}

\input{preamble}

\title{\LARGE \bf
Learning Agile Striker Skills for Humanoid Soccer Robots \\ from Noisy Sensory Input \vspace{-10pt}
}
\author{
  Zifan Xu$^{1}$, Myoungkyu Seo$^{1}$, Dongmyeong Lee$^{1}$, Hao Fu$^{1}$, Jiaheng Hu$^{1}$, Jiaxun Cui$^{1}$,\\ Yuqian Jiang$^{1}$, Zhihan Wang$^{1}$, Anastasiia Brund$^{1}$,
  Joydeep Biswas$^{1}$, Peter Stone$^{1,2}$%
  \\[0.6em]
  \small $^{1}$Department of Computer Science, The University of Texas at Austin, \small $^{2}$Sony AI\\
  \small {Project webiste: \tt\small \url{https://humanoidsoccer.github.io}}
}

\begin{document}
\twocolumn[{%
\renewcommand\twocolumn[1][]{#1}%
\maketitle

\begin{center}
\vspace{-20pt}
    {\includegraphics[width=0.85\textwidth]{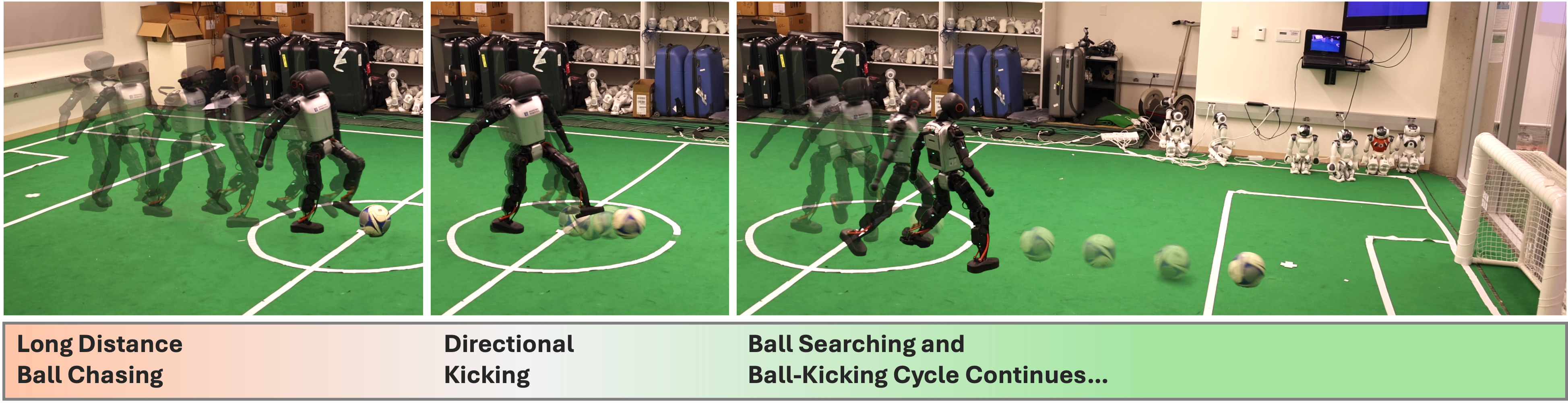}}
   \captionof{figure}{The illustration of a complete ball-kicking cycle in the robust continual ball-kicking task.
   }
\label{fig:teaser}
\end{center}%
}]

\makeatletter
\begingroup
\renewcommand\thefootnote{}%
\footnotetext{\@thanks This work has taken place jointly in the Learning Agents Research Group (LARG) and Autonomous Mobile Robotics Laboratory (AMRL) at UT Austin.  LARG research is supported in part by NSF (FAIN-2019844, NRT-2125858), ONR (N00014-24-1-2550), ARO (W911NF-17-2-0181, W911NF-23-2-0004, W911NF-25-1-0065), DARPA (Cooperative Agreement HR00112520004 on Ad Hoc Teamwork) Lockheed Martin, and UT Austin's Good Systems grand challenge. AMRL research is supported in part by NSF (CAREER-2046955, OIA-2219236, DGE-2125858, CCF-2319471), ARO (W911NF-23-2-0004), Amazon, and JP Morgan. Any opinions, findings, and conclusions expressed in this material are those of the authors and do not necessarily reflect the views of the sponsors. 
Peter Stone serves as the Chief Scientist of Sony AI and receives financial compensation for that role.  The terms of this arrangement have been reviewed and approved by the University of Texas at Austin in accordance with its policy on objectivity in research.
}%
\endgroup
\makeatother

\begin{abstract}
Learning fast and robust ball-kicking skills is a critical capability for humanoid soccer robots, yet it remains a challenging problem due to the need for rapid leg swings, postural stability on a single support foot, and robustness under noisy sensory input and external perturbations (e.g., opponents). This paper presents a reinforcement learning (RL)–based training pipeline that enables humanoid robots to execute robust continual ball-kicking with adaptability to different ball-goal configurations. The pipeline extends a typical teacher-student training framework---in which a ``teacher"  policy is trained with ground truth state information and the ``student" learns to mimic it with noisy, imperfect sensing---by including four training stages: (1) long-distance ball chasing (teacher); (2) directional kicking (teacher); (3) teacher policy distillation (student), and (4) student adaptation and refinement (student). Key design elements---including tailored reward functions, realistic noise modeling, and online constrained RL for adaptation and refinement---are critical for closing the sim-to-real gap and sustaining performance under perceptual uncertainty. Extensive evaluations in both simulation and on a real robot demonstrate strong kicking accuracy and goal-scoring success across diverse ball–goal configurations. Ablation studies further highlight the necessity of the constrained RL, noise modeling, and the adaptation stage. This work presents a training pipeline for robust continual humanoid ball-kicking under imperfect perception, establishing a benchmark task for visuomotor skill learning in humanoid whole-body control.
\end{abstract}

\input{content/intro.tex}
\input{content/related.tex}
\input{content/method.tex}
\input{content/deploy.tex}
\input{content/results.tex}

\section{Conclusions}
This paper introduced a RL–based system for enabling humanoid robots to perform fast and robust ball-kicking under noisy perception. Building on a four-stage teacher–student framework, comprising long-distance chasing, directional kicking, teacher policy distillation, and student adaptation and refinement, the system achieves continuous ball-kicking behaviors that remain stable under perceptual noise. Experiments in both simulation and on real hardware demonstrate high accuracy and goal-scoring success across diverse ball–goal configurations, while ablation studies confirm the necessity of core design choices including the noise modeling, constrained RL, and online adaptation.

A key limitation lies in that the long-distance chasing gaits rely on heavily engineered reward functions adapted from walking gait training, and the success of transferring to ball-kicking depends on their similarity in motion. This approach may not generalize to tasks with different motions, motivating future work on learning from diverse human motions as a more scalable strategy.

\bibliographystyle{IEEEtran}
\bibliography{IEEEabrv,references}

\end{document}

%% file: preamble.tex
\usepackage[dvipsnames]{xcolor}
\usepackage{mathtools}
\usepackage{booktabs}
\usepackage{amssymb}

\usepackage{graphicx}
\usepackage{caption} 
\usepackage{footnote}
\usepackage{hyperref}

\makesavenoteenv{table}

\newcount\Comments  
\newcommand{\zifan}[1]{{\ifnum\Comments=1\textcolor{blue}{[zifan: #1]}\fi}}
\newcommand{\commentp}[1]{{\ifnum\Comments=1\textcolor{red}{[Peter: #1]}\fi}}
\newcommand{\jb}[1]{{\ifnum\Comments=1\textcolor{red}{Joydeep:~#1}\fi}}
\newcommand{\jiaxun}[1]{{\ifnum\Comments=1\textcolor{orange}{Jiaxun:~#1}\fi}}

\definecolor{darkgreen}{RGB}{0,120,60}

\makeatletter
\newcommand{\printthanks}{\footnotetext[0]{\@thanks}}
\makeatother

%% file: content/intro.tex
\section{Introduction}
Humanoid robots are designed with anthropomorphic morphology, allowing them to operate in environments built for humans \cite{gu2025humanoid}. This makes them suitable for diverse tasks that potentially requires whole-body control~\cite{dao2024sim}, dynamic balance~\cite{li2025learning, zhang2025hub}, and coordinated interaction with the environment~\cite{wang2025end, li2025amo}. Recent advances in reinforcement learning (RL) and simulation have enabled progress in humanoid locomotion and manipulation, including robust locomotion~\cite{rudin2022learning, zhuang2023robot, agarwal2023legged}, and object interaction skills~\cite{he2024learning, barreiros2025careful, dadiotis2025dynamic}.
\commentp{It would be good to have 1 sentence at the start of the abstract that summarizes this paragraph before going into robot soccer.}

Despite this progress, learning whole-body tasks that combine speed, balance, and perception remains challenging. Such tasks require rapid limb movements, stability on a reduced support polygon, and robustness to perceptual noise and external perturbations. A prime example is the ball-kicking skill: the robot must approach the ball quickly, coordinate a powerful leg swing, maintain balance on one foot, and direct the ball toward the goal with precision. Effective striking further depends on accurate localization of both ball and goal, which is often degraded by sensor noise, delays, and perception errors. These demands make humanoid striking a difficult benchmark for visuomotor whole-body control.

This paper presents an RL–based training pipeline that enables humanoid robots to acquire such a reliable ball-kicking skill with great generalization across diverse ball–goal configurations. The pipeline comprises four consecutive training stages: (1) long-distance ball chasing, (2) directional kicking, (3) teacher policy distillation, and (4) student policy adaptation and refinement.
The first two training stages form a simple two-stage curriculum that enables a \emph{privileged teacher}, that has the ground-true ball and goal position information, to acquire ball-chasing skills, and extend them to directional kicking. During this process, aggressive domain randomization, such as external pushes on the body and ball, is applied to encourage recovery from imperfect states, such as missed kicks and tilted postures. The teacher policy is then distilled into a student policy, that has \emph{imperfect} perception of the ball and goal position, via DAgger~\cite{ross2011reduction}. The modeling of such imperfect perception constitutes three components: a velocity-dependent noise model, delayed updates, and frame drops caused by perceptual occlusion. Finally, stage four applies online adaptation using N-P3O~\cite{lee2023evaluation}, a constrained RL algorithm, to refine the kicking skill. This process reduces jittery leg motions and unsafe sharp turning that often arise when using a fixed regularization coefficient, which can impose disproportionately large penalties at certain steps—particularly those immediately preceding the kick, where the agent expects high immediate future reward from the kicking task.

We evaluate the resulting policy extensively in simulation, measuring kick accuracy, success rate, kick strength, and energy cost across diverse ball–goal configurations. Ablation studies highlight the importance of the final adaptation stage and motion refinement through constrained RL. To assess real-world performance, we deploy the policy on a Booster T1 humanoid robot, which achieves an average success rate of 66.7\% across five different ball–goal configurations. Overall, this work presents a comprehensive framework for learning robust continual humanoid ball-kicking skill under imperfect perception and establishes a benchmark for agile visuomotor skill learning in humanoid robots.

%% file: content/related.tex
\section{Related Work}

\subsection{Robot Soccer}
Robot soccer has long served as a benchmark domain for testing integrated perception, control, and decision-making systems \cite{kohl2004policy,hausknecht2010learning,gerndt2015humanoid}. Early efforts explored rule-based strategies and simple controllers, but recent advances in reinforcement learning (RL) have enabled agile, dynamic skills. On quadrupeds, researchers have demonstrated soccer behaviors such as cooperative play \cite{SuEtAl2025_QuadrupedalSoccer} and precise shooting \cite{DribbleBot2023_JiMargolisAgrawal} using hierarchical RL, robust dribbling in the wild \cite{Ji2022_HierRL_SoccerKicking}, and dynamic goalkeeping against human players~\cite{Blommers2024_QuadrupedalGoalkeeping, Huang2023_QuadrupedGoalkeeper}. Quadrupeds naturally benefit from high stability, making perception and balance control easier compared to humanoid platforms.

Humanoid soccer is comparatively less explored, but it presents greater challenges due to bipedal balance, whole-body coordination, and larger perceptual uncertainty~\cite{HaarnojaEtAl2024_LearningAgileSoccerSkills, wang2025learning}. Prior work has focused on dribbling with agile locomotion~\cite{Wang2025_DribbleMaster}, learning from gameplay recordings, and diffusion-based visuomotor policies~\cite{Vahl2025_SoccerDiffusion}. Other efforts target biomechanics-inspired approaches to humanoid kicking~\cite{MarewEtAl2024_BiomechanicsKick}. However, most prior humanoid studies emphasize dribbling or generic locomotion skills, while agile striking under noisy perception remains relatively unaddressed. Our work fills this gap by tackling robust and adaptive ball-kicking skills, a critical capability for advancing humanoid soccer.

\subsection{Humanoid Whole-Body Control}
Whole-body control of humanoid robots is a central challenge in robotics, requiring dynamic coordination of dozens of degrees of freedom while maintaining balance and stability~\cite{kim2016dynamic, sentis2006whole, he2025asap, cheng2024expressive, hu2023causal, ferigo2021emergence, hu2025slac, ma2025learning, zhang2024wococo}. Traditional methods often rely on model predictive control (MPC) or optimization-based controllers that explicitly encode dynamics and contact constraints~\cite{kim2016dynamic, sentis2006whole}. While these approaches provide stability guarantees, they can be brittle under perception noise, latency, or unmodeled dynamics.

Reinforcement learning has recently emerged as a powerful alternative, enabling humanoids to learn locomotion~\cite{li2021reinforcement}, acrobatic maneuvers~\cite{he2025asap}, and robust recovery strategies~\cite{ferigo2021emergence} directly from interaction. GPU-accelerated simulation frameworks have further accelerated progress, allowing large-scale training of bipedal policies. Beyond locomotion, researchers have begun to study whole-body visuomotor skills, such as manipulation~\cite{hu2025slac}, agile sports behaviors (e.g., badminton~\cite{ma2025learning}), and multi-contact motion tracking~\cite{zhang2024wococo}. These advances highlight the feasibility of training policies that integrate perception and control for high-dimensional humanoids.

Our work contributes to this growing literature by extending whole-body control to a challenging soccer striker task. Unlike locomotion or manipulation, striking requires rapid leg swings, balance on a single support foot, and accurate timing under noisy perception. By combining robust perception with a staged RL framework, our approach bridges the gap between whole-body control research and the practical demands of humanoid soccer.

%% file: content/method.tex
\section{Method}
\begin{figure*}
    \centering
    \includegraphics[width=\linewidth]{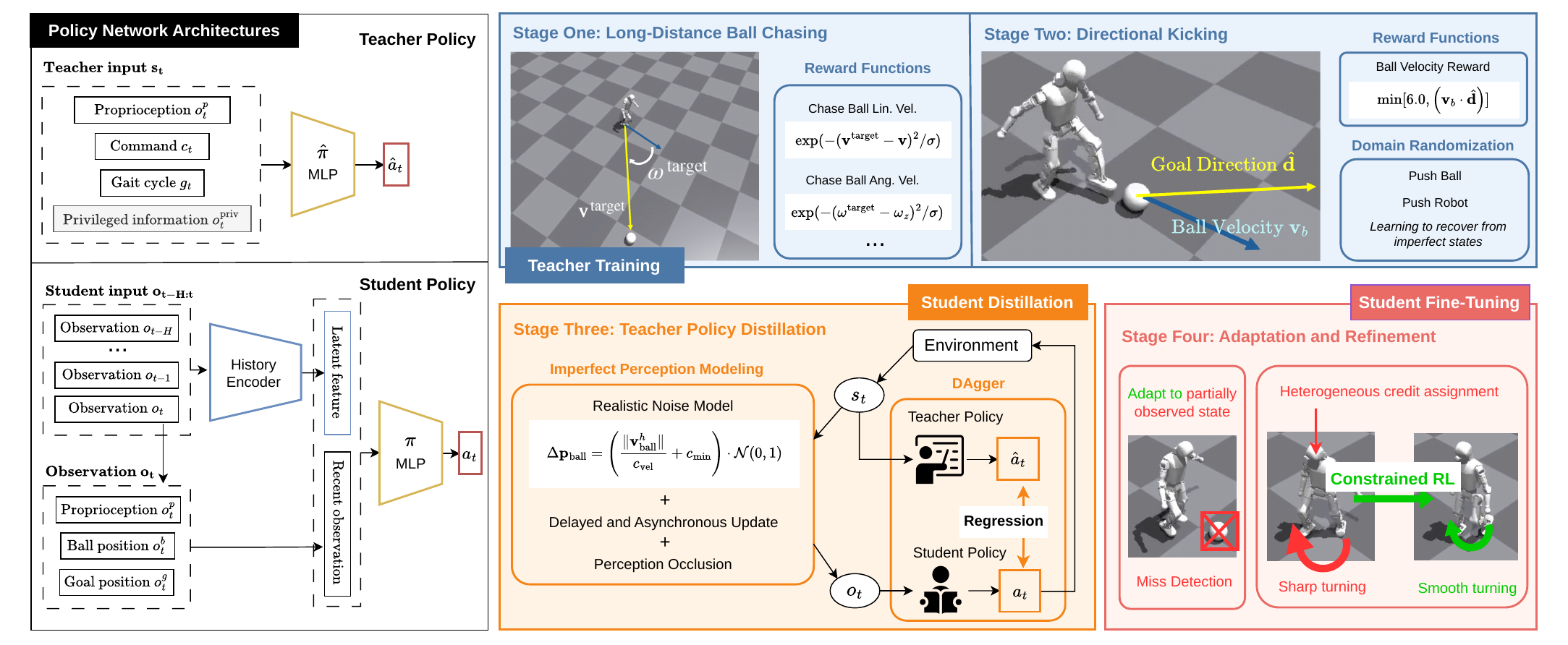}
    \caption{Left: The network architectures for the teacher and the student network; Right: Multi-stage training framework: (1) Long-distance Chasing; (2) Directional Kicking; (3) Teacher Policy Distillation; (4) Student Adaptation and Refinement.
    }
    \vspace{-10pt}
    \label{fig:system }
\end{figure*}
\begin{figure}
    \centering
    \includegraphics[width=0.9\linewidth]{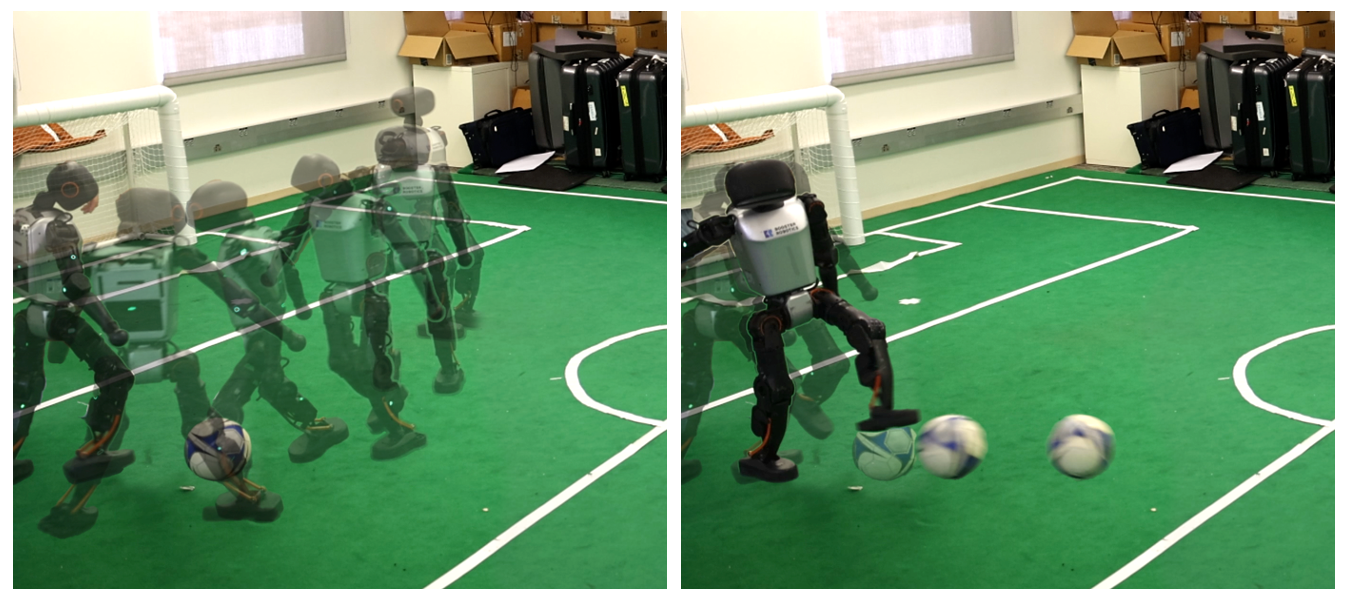}
    \caption{Kicking cycle phase (iii): reorienting to locate the ball and seamlessly initiating a kicking attempt.}
    \label{fig:placeholder}
    \label{fig:reorientation}
    \vspace{-10pt}
\end{figure}

\textbf{System Overview.}  
The objective of the system is to learn a control policy $a_t = \pi(o_{t-H:t})$ that maps a history of \(H\) observations \(o_{t-H:t} = [o_{t-H}, \ldots, o_t]\) to an action \(a_t\), corresponding to the joint position targets of a humanoid robot at 50~Hz. Each observation at time \(t\) is a tuple $o_t = (o^p_t, o^b_t, o^g_t)$ where \(o^p_t\) denotes proprioceptive measurements, and \(o^b_t\) and \(o^g_t\) represent the estimated ball and goal positions, respectively. By executing the learned policy, the robot exhibits continual ball-kicking behaviors. As shown in Fig.~\ref{fig:teaser}, each kicking cycle integrates three key phases: (i) approaching the ball from long distance, (ii) performing a kick motion that directs the ball toward the goal, and (iii) reorienting to locate the ball again and seamlessly initiating the next kicking attempt. An illustration of phase (iii) is shown in Fig.~\ref{fig:reorientation}.
\commentp{None of the videos you showed us in the real world actually do phase  (iii).  Can we actually do that in the real world?}

The system is trained through a four-stage training framework, followed by a real-world deployment pipeline:
\begin{enumerate}
    \item \textbf{Long-Distance Chasing} -- a ``teacher" policy, by which we mean a policy  with privileged ground-truth ball position, learning a robust walking gait to approach the ball from diverse initial configurations; 
    \item \textbf{Directional Kicking} -- a teacher policy acquiring precise and robust kick motions using privileged ground-truth ball and goal positions;  
    \item \textbf{Teacher Policy Distillation} -- transferring privileged-information teacher policies into ``student" policies, by which we mean policies that operate under imperfect perception.  
    \item \textbf{Student Adaptation and Refinement} -- applying an online constrained RL algorithm to adapt to partially observed states created by imperfect perception.
\end{enumerate}

Each stage builds on the previous one, progressively shaping the policy to acquire a ball-kicking skill that is robust to perception noise enabling zero-shot real-world deployment. The following sections detail the four training stages and the deployment pipeline.

\subsection{Long-Distance Chasing}
\paragraph{Privileged Teacher Policy} The teacher policy is defined as $\hat{a}_t = \hat{\pi}(s_t)$ where the input state $s_t$ at timestep $t$ consists of three components: (i) proprioceptive observations $o^p_t$, (ii) a velocity command $c_t = (v^x_t, v^y_t, \omega^z_t$), (iii) a gait cycle $g_t$, and (iv) privileged observations $o^{\text{priv}}_t$. The design of the proprioceptive and privileged observations varies across the actual morphology of the robot. In this paper, we follow the proprioceptive and privileged observations from Booster Gym~\cite{wang2025booster}, with the exception that the privileged input additionally includes ground-truth information about the ball and goal positions, the ball’s linear velocity, and its physical parameters such as mass and friction. The full list of observations is listed in Table~\ref{tab:obs}.
\begin{table}[t]
\centering
\caption{Summary of Observation Space.}
\label{tab:obs}
\resizebox{0.8\linewidth}{!}{%
\begin{tabular}{l c c c c}
\toprule
\textbf{Components} & \textbf{Dims} & \textbf{Teacher (Critic)} & \textbf{Student} \\
\midrule
Commands          & 3  & \checkmark & \checkmark \\
Gait Cycle        & 2  & \checkmark & \checkmark \\
Gravity Vector    & 3  & \checkmark & \checkmark \\
Angular Velocity  & 3  & \checkmark & \checkmark \\
Joint Position    & 23 & \checkmark & \checkmark \\
Joint Velocity    & 23 & \checkmark & \checkmark \\
Previous Action   & 23 & \checkmark & \checkmark \\
Noisy Ball Position & 2 &  & \checkmark & \\
Noisy Goal Position & 2 &  & \checkmark & \\
\midrule
Body Mass         & 1  &      \checkmark      & \\
Body Center of Mass & 3 &     \checkmark      & \\
Base Linear Velocity & 3 &    \checkmark      & \\
Base Height       & 1  &      \checkmark      & \\
Push Force        & 2  &      \checkmark     & \\
Push Torque       & 3  &      \checkmark      & \\
Real Ball Position & 2  &    \checkmark        & \\
Real Goal Position & 2  &    \checkmark        & \\
Real Ball Velocity & 2  &    \checkmark        & \\
Ball Mass & 1  &    \checkmark        & \\
Ball Friction & 1  &    \checkmark        & \\
\bottomrule
\end{tabular}
}
\vspace{-10pt}
\end{table}

\paragraph{Environment Design} The environment contains a humanoid robot, a soccer ball, and a point goal representing the kick direction. At the beginning of each episode, the robot’s initial position is sampled uniformly within an annular region centered at the ball, bounded by an inner radius of 0.5 m and an outer radius of 4.0 m. The long-distance chasing task is adapted from the standard flat-ground walking task, with modifications tailored to the ball-chasing task. Instead of tracking a velocity target in the world frame, the commanded linear velocity always aligns with the robot–ball vector. Let $\mathbf{r}_{rb}(t) \in \mathbb{R}^2$ denote the vector from the robot base to the ball at time $t$, and $\hat{\mathbf{r}}_{rb}(t) = \tfrac{\mathbf{r}_{rb}(t)}{|\mathbf{r}_{rb}(t)|}$ its unit vector. The commanded linear velocity target is defined as $\mathbf{v}_t^{target} = v_t^x \cdot \hat{\mathbf{r}}_{rb}(t)$. For rotation, let $\theta_t \in [-\pi, \pi]$ be the signed angular difference between the robot heading vector and $\hat{\mathbf{r}}_{rb}(t)$. The target angular velocity is given by $\omega_t^{target}=\omega_t^{z} \cdot \text{sign}(\theta_t)$, so that the robot always turns in the direction (left or right) that minimizes $|\theta_t|$. The scalar $\omega^z_t \geq 0$ controls the magnitude of the turning speed.

To improve robustness, we apply domain randomization following prior work on humanoid locomotion~\cite{wang2025booster,unitree_rl_gym}. Randomized parameters include joint stiffness, damping, and friction, body masses, and center of mass, as well as external perturbations such as forces and pushes. Additionally, the ball’s mass and surface friction are randomized to encourage robust ball interaction. 

\paragraph{Reward Design} The reward formulation builds on prior walking-gait designs~\cite{wang2025booster,unitree_rl_gym}, where the primary task rewards are exponential kernels of linear and angular velocity tracking errors. For the ball-chasing task, the tracking errors are defined relative to $\mathbf{v}_t^{target}$ and $\omega_t^{target}$ grounded by the ball position (detailed in Section III-B (a)). Table~\ref{tab:rewards} provides the full list of reward functions specific to our hardware platform Booster T1 robot.

\begin{table}[!t]
\centering
\resizebox{0.8\linewidth}{!}{%
\begin{tabular}{lcc}
\toprule
\textbf{Components} &\textbf{Equations}\footnote{The notation in this Table is consistent with popular convention, and will be defined in full detail in the Appendix of the online publication.} & \textbf{Weights} \\
\midrule
\multicolumn{3}{c}{\text{Task Rewards}} \\
\midrule
Chase ball lin. vel. & $\exp(-(\mathbf{v}^{\text{target}}-\mathbf{v})^2/\sigma)$ & 2.0 (0.5) \\
Chase ball ang. vel. &$\exp(-(\omega^{\text{target}}-\omega_z)^2/\sigma)$ & 2.0 (0.5) \\
Feet swing & $1_{\text{feet swing}}\cdot1_{\text{swing period}}$ & $3.0$ (0.0) \\
Arm swing & $1_{\text{arm swing}}\cdot1_{\text{swing period}}$ & 1.0 (0.0) \\
Head tracking ball & $1_{\text{ball in FOV}}$ & 1.0 \\
Ball velocity & $\min[6.0, \left(\mathbf{v}_b \cdot \hat{\mathbf{d}}\right)]$ & 0.0 (4.0) \\
\midrule
\multicolumn{3}{c}{Regularization Rewards} \\
\midrule
Survival & 1 & 1.0 \\
Base height & $(h^{\text{des}}-h)^2$ & $-20.0$ \\
Orientation & $\|\boldsymbol{g}\|^2$ & $-5.0$ \\
Torque & $\|\boldsymbol{\tau}\|^2$ & $-2\times 10^{-4}$ \\
Torque tiredness & $\|\boldsymbol{\tau}/\boldsymbol{\tau}_{\max}\|^2$ & $-1\times 10^{-2}$ \\
Power & $\max(\boldsymbol{\tau}\cdot\dot{\boldsymbol{q}},0)$ & $-2\times 10^{-4}$\\
Lin velocity (z) & $v_z^2$ & $-2.0$ \\
Ang velocity (xy) & $\|\boldsymbol{\omega}_{xy}\|^2$ & $-0.2$ \\
Joint velocity & $\|\dot{\boldsymbol{q}}\|^2$ & $-1\times 10^{-4}$ \\
Joint acceleration & $\|\ddot{\boldsymbol{q}}\|^2$ & $-1\times 10^{-7}$ \\
Base acceleration & $\|\dot{\boldsymbol{v}}\|^2+\|\dot{\boldsymbol{\omega}}\|^2$ & $-1\times 10^{-4}$ \\
Action rate & $\|\boldsymbol{a}_t-\boldsymbol{a}_{t-1}\|^2$ & $-1.0$ \\
Upper body symmetry & $\| q - q_{0} \|^2$ & $-0.5$ \\
Joint position limit & $1_{\boldsymbol{q}>\boldsymbol{q}_{\max}}+1_{\boldsymbol{q}<\boldsymbol{q}_{\min}}$ & $-1.0$ \\
Collision & $n_{\text{collision}}$ & $-1.0$ \\
Feet slip & $1_{\text{feet stance}}\cdot \|\boldsymbol{v}_{\text{feet}}\|^2$ & $-0.1$ \\
Feet yaw & $\|\boldsymbol{\psi}_{\text{feet}}-\psi_{\text{base}}\|^2$ & $-1.0$ (0.0)  \\
Feet roll & $\|\boldsymbol{\phi}_{\text{feet}}\|^2$ & $-0.1$  \\
Feet distance & $\max(d_{\text{ref}}-d_{\text{feet}},0)$ & $-1.0$  \\
\bottomrule
\end{tabular}
}
\caption{Summary of reward functions. Weights are specified for the long-distance chasing task; values in parentheses indicate the weights used for the directional kicking task when they differ.}
\label{tab:rewards}
\vspace{-20pt}
\end{table}

\subsection{Directional Kicking}
\paragraph{Environment Design}
The directional kick environment simulates a soccer field consistent with the RoboCup Adult Size Humanoid League rules \cite{RoboCupHLRules2025}. The field measures $14  \text{ m} \times 9 \text{ m}$, with goals placed at $x=-7 \text{ m}$ and $x=+7 \text{ m}$.

At the beginning of each episode, the ball is uniformly initialized within the region $[-1.5, 6.5] \times [-4.0, 4.0]$, corresponding roughly to the front half of the field. The robot is then placed within an annular region centered at the ball, bounded by an inner radius of $0.5 \text{ m}$ and an outer radius of $2.0 \text{ m}$.

If the ball exits the field boundary, it is reset to a new position without terminating the episode. The reset location is sampled from an annular region centered at the original ball position. This setup encourages the robot to repeatedly search for the ball, approach it, and execute a kick, thereby creating continuous striking cycles.

To improve robustness, we introduce additional domain randomization: every four seconds, the ball receives an external disturbance by applying a random linear velocity sampled from $[-0.1, 0.1] \text{ m/s}$ along both axes. This perturbation produces imperfect states from which the teacher policy must recover. These recovery behaviors can then be distilled into the student policy, which may encounter similar imperfect states due to noisy perception.

\paragraph{Reward Design} The directional kick task introduces two extra reward functions: \emph{ball velocity} reward and \emph{head tracking ball} reward detailed as follows.

\emph{Ball velocity} reward encourages the robot to impart a high ball velocity directed toward the opponent’s goal. Let $\mathbf{p}_g \in \mathbb{R}^2$ denote the goal position, $\mathbf{p}_b \in \mathbb{R}^2$ the ball position, and $\mathbf{v}_b \in \mathbb{R}^2$ the linear velocity of the ball. Define the goal direction unit vector as
\begin{equation}
\hat{\mathbf{d}} = \frac{\mathbf{p}_g - \mathbf{p}_b}{\|\mathbf{p}_g - \mathbf{p}_b\|}.
\end{equation}
The reward is given by the projection of the ball velocity onto the goal direction, scaled by the velocity magnitude:
\begin{equation}
r_t = \min[6.0, \left(\mathbf{v}_b \cdot \hat{\mathbf{d}}\right)],
\end{equation}
The reward is capped at 6.0 to prevent the agent from exploiting the objective by generating excessively large ball velocities.

\emph{Head tracking ball} reward encourages the robot to maintain visual contact with the ball. At each timestep, a reward of $+1$ is assigned if the ball lies within the field of view of the robot’s head camera.

\subsection{Teacher Policy Distillation}
Once the teacher policy acquires a basic kicking skill with a degree of recovery capability, it is distilled into a student policy that can operate under imperfect perception.

\paragraph{Student Policy} The student policy is defined as $a_t = \pi(o_{t-H:t})$, where the input consists of a history of $H=50$ observations $o_{t-H:t} = [o_{t-H}, \ldots, o_t]$. Each observation is represented as $o_t = (o^p_t, o^b_t, o^g_t)$, with $o^p_t$ denoting proprioceptive measurements, and $o^b_t$ and $o^g_t$ corresponding to noisy estimates of the ball and goal positions. The perceptual error modeling for these inputs is described in the next subsection.

To capture temporal dependencies, the observation history $o_{t-H:t}$ is first processed by a 1D convolutional encoder, producing a 64-dimensional latent representation. This latent vector is concatenated with the most recent observation $o_t$ and passed through a multilayer perceptron (MLP) with three hidden layers of 256, 256, and 128 units, respectively, to output the action.

\paragraph{Noise Modeling} To simulate realistic perception uncertainty, we employ a velocity-dependent noise model~\cite{li2025clone} for both the ball and goal positions. The magnitude of the injected noise scales with object velocity in the robot's head frame, reflecting the fact that rapid-moving objects generally lead to larger noise in camera-based localization. Formally, the noise to the ball/goal position $\Delta \mathbf{p}_{\text{ball/goal}}$ is defined as
\begin{equation}
\Delta \mathbf{p}_{\text{ball/goal}} = \left( \frac{\|{\mathbf{v}}^h_{\text{ball/goal}}\|}{c_{\text{vel}}} + c_{\min} \right) \cdot \mathcal{N}(0,I),
\end{equation}
where ${\mathbf{v}}^h_{\text{ball/goal}}$ denotes the ball/goal velocity in the robot's camera frame, $c_{\text{vel}}$ and $c_{\min}$ are scaling constants, and $\mathcal{N}(0,I)$ is standard Gaussian vector noise. The student policy is trained using Dataset Aggregation (DAgger)~\cite{ross2011reduction}, which iteratively collects rollouts under the student policy while querying the teacher for corrective actions. This process ensures that the student learns not only from the teacher’s demonstrations but also from states encountered due to its own errors, improving robustness under noisy perception.

\begin{figure}
    \centering
    \includegraphics[width=\linewidth]{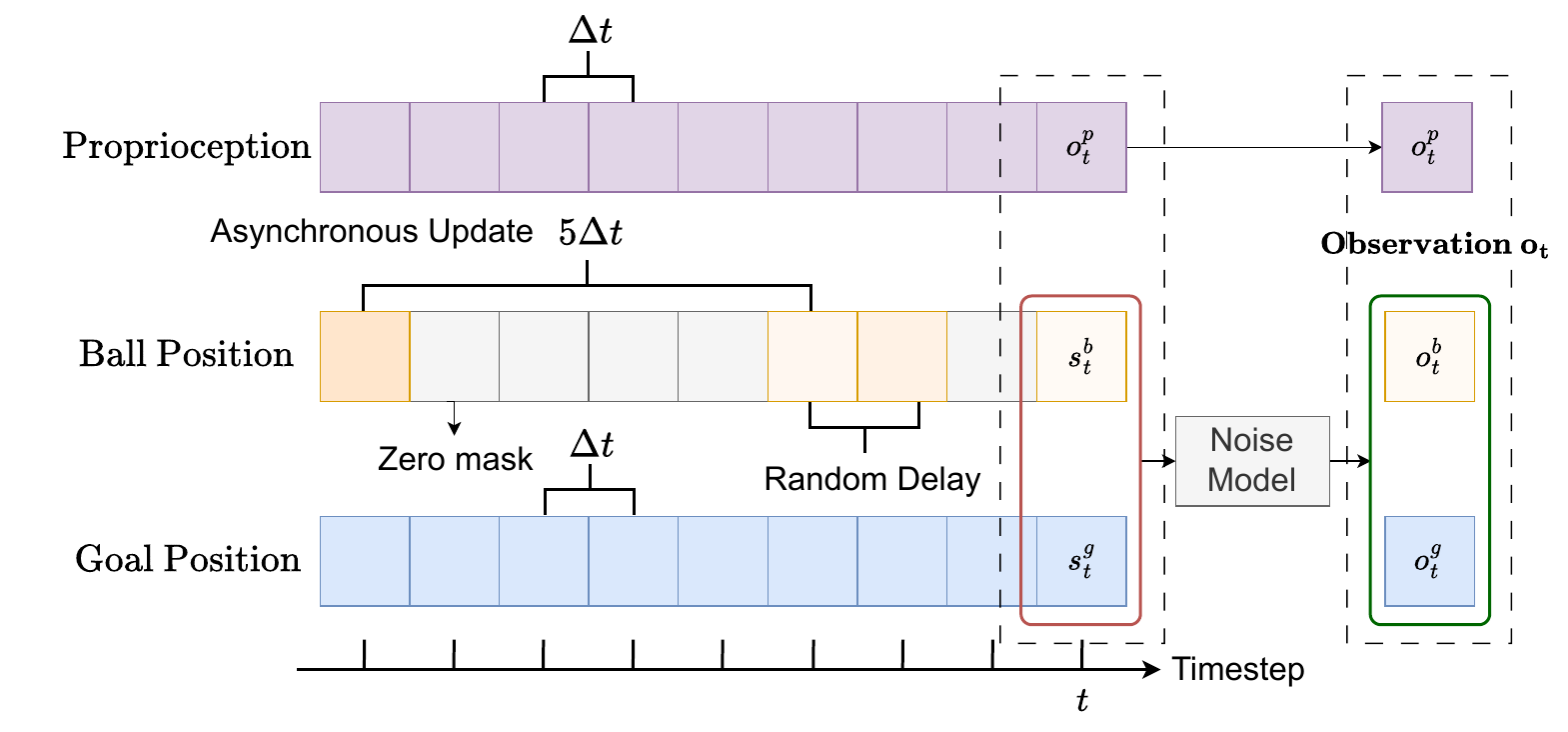}
    \caption{An illustration of the realistic perception modeling. 
    }
    \vspace{-10pt}
    \label{fig:noise}
\end{figure}
In addition, perception signals are subject to realistic delays and dropouts. Proprioceptive observations are updated at 50 Hz (every 0.02 s), while ball positions are published at around 10 Hz (every 0.1 s) with a random offset to mimic asynchronous updates. If the ball lies outside the camera’s field of view or falls between publish intervals, its position $o^b_t$ is masked with zeros. Goal observations are updated at the same frequency as the proprioception and subject to the same noise injection. 
Fig. \ref{fig:noise} shows an overview of the imperfect perception model.

\subsection{Student Adaptation and Refinement}
\paragraph{Student Adaptation} Conventional teacher–student frameworks typically involve training a teacher policy and distilling it into a student policy~\cite{agarwal2023legged}, under the assumption that full environment state information can be reconstructed from observation histories. However, recent work shows that certain partially observed states cannot be effectively captured by the teacher training stage, and that online RL-based adaptation can significantly improve performance in such cases~\cite{zhang2025distillation}. Inspired by this work, we further adapt the student policy with online RL under noisy perception. During adaptation, we apply smaller perturbations to the robot and ball, allowing the policy to concentrate on improving accuracy and robustness to realistic sensory uncertainty.

\begin{figure}
    \centering
    \includegraphics[width=0.85\linewidth]{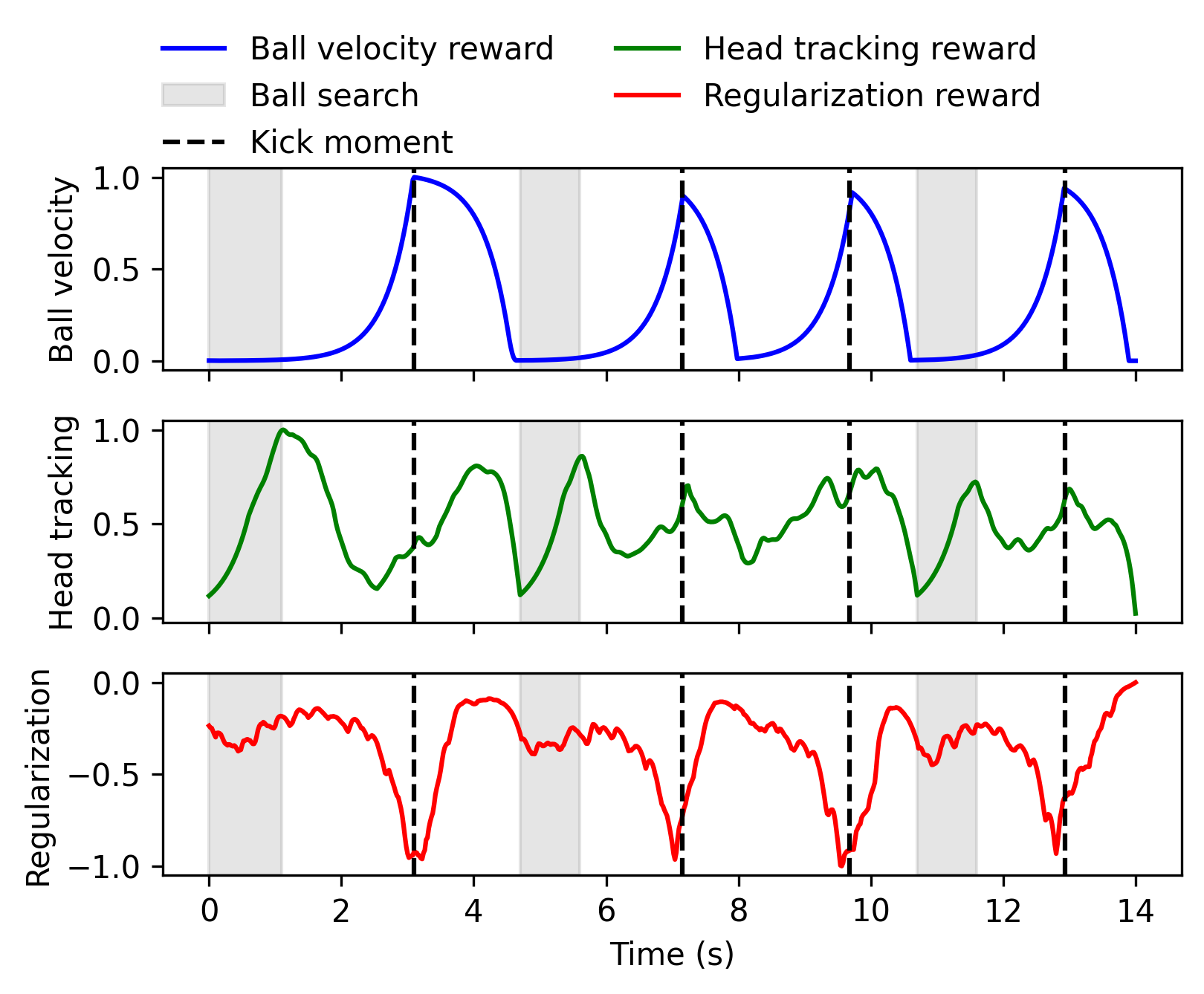}
    \caption{An illustration of inhomogeneous credit assignment within four complete kicking cycles. y-axis shows the normalized future returns for regularization reward, head tracking reward, and ball velocity reward, respectively. Large dips in regularization reward indicate jittery kick motions. 
    }
    \label{fig:heterogeneous}
    \vspace{-20pt}
\end{figure}

\paragraph{Refinement through constrained RL} While training the adapted student, we observed undesirable behaviors such as sharp rotations during ball search and overly dynamic kicking, which may compromise both safety and stability during the real-world deployment. We attribute these issues to inhomogeneous credit assignment within an episode, where task rewards, e.g., \emph{ball velocity} in Fig.~\ref{fig:heterogeneous} (\textcolor{blue}{top}) and \emph{head tracking ball} in Fig.~\ref{fig:heterogeneous} (\textcolor{darkgreen}{middle}), are distributed unevenly across a ball-kicking cycle. In particular, the discounted future reward associated with ball velocity exhibits a pronounced peak at the kick moment (indicated by the dashed vertical lines in Fig.~\ref{fig:heterogeneous}). In this case, a fixed regularization coefficient can impose disproportionately large penalties at certain steps—particularly those immediately preceding the kick, where the agent expects high immediate future reward from the kicking task. This effect is reflected in the sharp drops of the regularization reward shown in Fig.~\ref{fig:heterogeneous} (\textcolor{red}{bottom}).

To mitigate this problem, we reformulate directional kicking as a constrained RL problem. Let $r_{\text{task}}$ denote the aggregated task rewards and $r_{\text{regu}}$ denote the aggregated regularization rewards (as defined in Table~\ref{tab:rewards}). The objective is to maximize task performance while constraining excessive regularization costs:
\begin{align}
\max_{\pi} & \quad J_{\text{task}}(\pi) \\
\text{s.t.} & \quad r_{\text{regu}}(a_t, s_t) \leq h, \quad \forall (s_t, a_t) \in \tau, ; \tau \sim p^\pi(\tau),
\end{align}
where $J_{\text{task}}(\pi) = \mathbb{E} \Bigl[\sum_{t=0}^T \gamma^t r_{\text{task}}(a_t, s_t) \mid s_0 \sim \mu(s) , \tau \sim p^\pi(\tau)\Bigr]$ 
with $\mu(s)$ denotes the initial state distribution, $\tau$ a trajectory sampled under policy $\pi$, and $h$ is the upper limit on the aggregated regularization reward. In practice, $h$ is set to $1.5\times$ per-step average regularization cost of a walking gait.

This constrained optimization is solved using N-P3O \cite{lee2023evaluation}, a constrained policy gradient method shown to be effective for locomotion tasks \cite{ma2025learning}. Importantly, the role of constrained RL here is not strict hardware safety enforcement, but rather the mitigation of inhomogeneous credit assignment, resulting in smoother, and safer ball-kicking motions.

%% file: content/deploy.tex
\section{Real-World Deployment}
This section details the hardware platforms and core components that enables the real-world deployments.
\begin{figure}
    \centering
    \includegraphics[width=\linewidth]{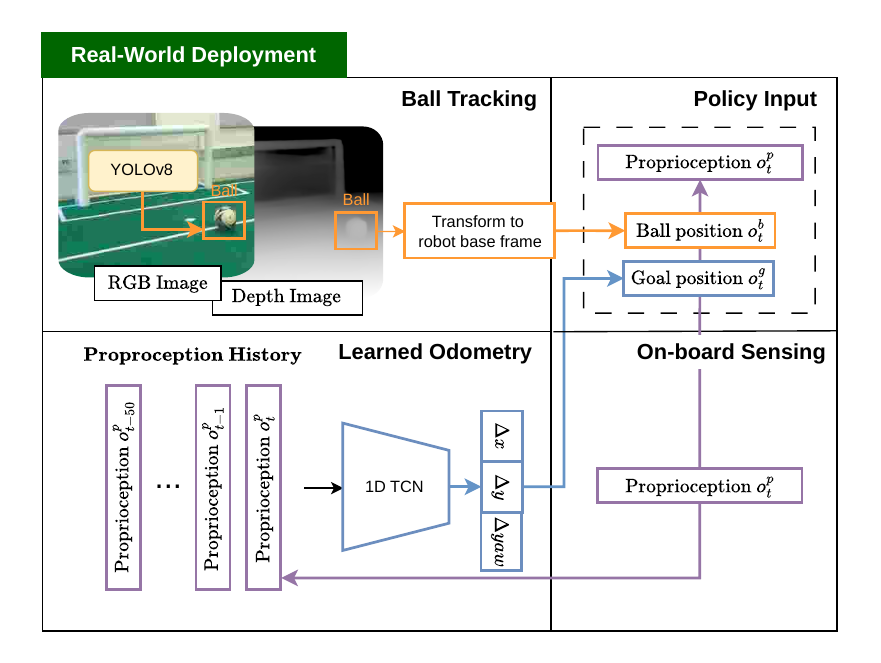}
    \caption{An overview of the real-world deployment pipeline}
    \label{fig:deployment}
    \vspace{-20pt}
\end{figure}

\subsection{Hardware Platform} We deploy our ball-kicking policy on the physical Booster T1 humanoid robot~\cite{BoosterRobotics_T1}. The robot stands 1.18-meter-tall, features 23 degrees of freedom and is equipped with a rich sensor suite including an IMU and joint encoders for real-time state estimation, as well as a ZED 2i Stereo Camera for visual perception. All onboard computations are performed using an NVIDIA AGX Orin GPU in conjunction with a 14-core high-performance CPU. The ball-kicking policy is executed at a control frequency of 50 Hz.

\subsection{Perception Pipeline}
Our perception pipeline provides the policy with real-time estimates of the ball and the opponent's goal in the robot's base frame. For ball perception, a YOLOv8 model~\cite{varghese2024yolov8} processes the robot's RGB-D camera stream to detect the ball's bounding box. The center of the detected box is unprojected into a 3D point using its corresponding depth value. This point is then transformed into the robot's base frame using the head's forward kinematics. This results the pose of the ball in the robot's base frame serving as the input to the ball-kicking policy. 

\subsection{Odometry System}
To enable goal localization, we employ a lightweight data-driven legged-inertial odometry module inspired by Legolas~\cite{wasserman2024legolas}. As shown in Fig.~\ref{fig:deployment} on the bottom left, the system fuses inertial signals—specifically the gravity vector and angular velocity—with leg kinematics such as joint positions and velocities. A neural network processes a 50-frame history of these measurements and predicts the relative pose change $(\Delta x, \Delta y, \Delta \text{yaw})$ in SE(2) space. Robustness of such prediction is further enhanced through multi-step supervision and an SE(2) consistency constraint, which together promote long-term stability and coherent estimation of both translation and rotation. Across trajectories spanning four complete ball-kicking cycles, the learned odometry achieves an average position error of $0.14 \pm 0.09$ meters.



%% file: content/results.tex
\section{Evaluations}
\begin{figure*}[htb!]
    \centering
    \includegraphics[width=0.90\linewidth]{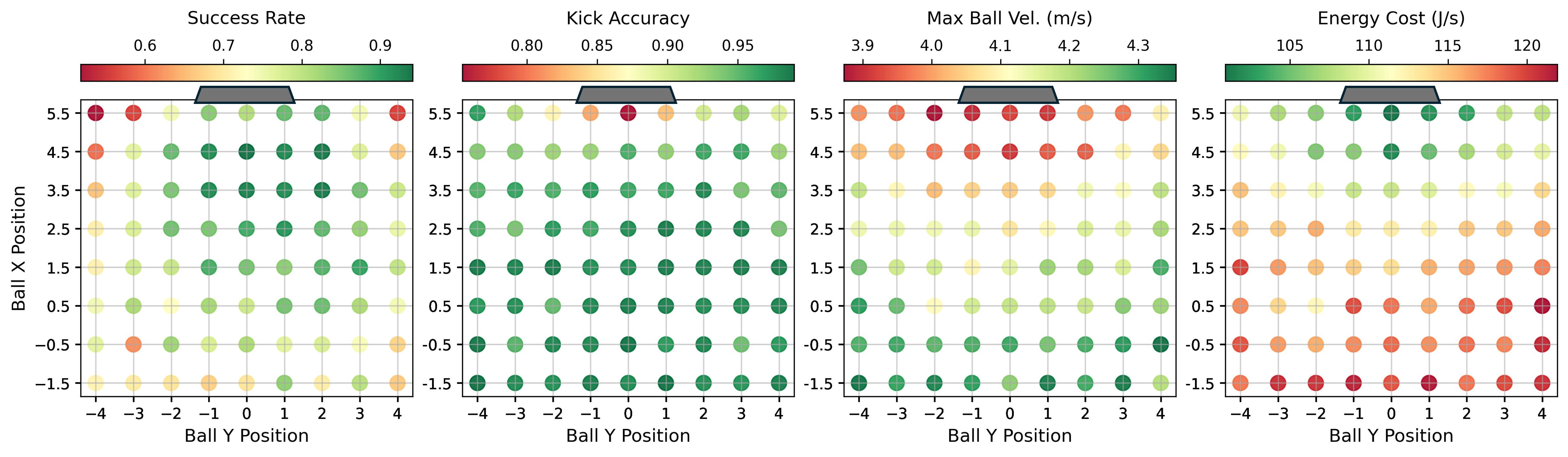}
    \caption{Visualizations of success rate, kick accuracy, max ball vel., and energy cost, at different initial ball positions. The shadowed trapezoidal regions visualize the goal areas.}
    \vspace{-10pt}
    \label{fig:success}
\end{figure*}
The learned control policy is evaluated in both simulation and the real world. We detail the evaluation settings and present the results in the following sections.

\subsection{Simulation Results}
\paragraph{Evaluation setting} We conduct evaluation under a diverse set of initial ball positions, represented as a $9 \times 9$ matrix evenly spanning the region $[-1.5, 6.5] \times [-4.0, 4.0]$ at 1 m intervals. For each ball position, 50 independent trials are executed with randomly sampled robot initial states.

\paragraph{Kick accuracy and success rate} Kick accuracy is measured as the cosine similarity between the post-kick ball velocity unit vector 
and the ground-truth ball–goal direction vector. The \emph{success rate} is defined as the fraction of trials in which the ball crosses the goal line between the the goal posts\footnote{The goal posts distance are specified by RoboCup Adult-Size Humanoid League rules~\cite{RoboCupHLRules2025}.}. Under these metrics, the policy achieves an average success rate of 79.5\% and an average kick accuracy of 0.956. Results are visualized in Fig.~\ref{fig:success}

\paragraph{Maximum ball velocity and energy cost} We also evaluate the maximum ball velocity and the average energy cost during kicking, where energy cost is computed over timesteps in which the robot is within $0.5 \text{ m}$ of the ball. The learned policy achieves an overall maximum ball velocity of $4.13 \text{ m/s}$ and an average energy cost of $110.8 \text{ J/s}$. Figure~\ref{fig:success} (right) illustrates the distribution of these metrics across different initial ball positions. Notably, an emergent energy-saving strategy is observed: when the ball is positioned closer to the goal, the robot consistently applies lighter kicks and expends less energy, indicating adaptive behavior that balances efficiency with task success.

\subsection{Real-world results} We perform a smaller-scale evaluation on hardware using five different ball positions in a RoboCup Kid-Size soccer field. For each position, three trials are conducted with the robot initialized from the same fixed location ($6.5 \text{ m}$ in front of the goal). We report averaged kick accuracies and success rates across these trials. The results are presented in Table.~\ref{tab:real}. The policy achieves an overall success rate of 66.7\%.
\begin{table}[bth!]
\centering
\begin{tabular}{cccc}
\toprule
Ball Pos. x (m) & Ball Pos. y (m) & Kick Accuracy & Success Rate \\
\midrule
4.5           & ~0.0           &       0.99 $\pm$ 0.01        &  3/3                  \\
3.5           & ~1.8          &       0.94 $\pm$ 0.08        &   2/3                 \\
3.5           & -1.8           &      0.95 $\pm$ 0.02        &   1/3                 \\
6.0           & ~2.5           &       0.99 $\pm$ 0.02        &  2/3                  \\
6.0           & -2.5          &       0.99 $\pm$ 0.01        &   2/3                 \\
\bottomrule
\end{tabular}
\caption{Kick accuracy and success rate for the real-world experiments at different ball positions (relative to goal center). The robot starts from 6.5 meters away from the goal.}
\label{tab:real}
\vspace{-20pt}
\end{table}

\subsection{Ablation Studies}
To understand the contribution of individual design components, we perform ablations on the two key aspects of our system:


\paragraph{Constrained RL}
We evaluate the effect of constrained RL by comparing N-P3O against standard PPO with fixed regularization coefficients. Table~\ref{tab:constrainedRL} summarizes the results across success rate, peak ball velocities, and energy cost. N-P3O achieves the highest overall performance, with a success rate of 79.5\%, while maintaining smooth and stable motions characterized by lower maximum angular velocity and energy consumption at a small cost to peak ball velocity. Energy consumption is also substantially reduced to $108.6 \text{ J/s}$, less than half that of PPO. These results confirm that the constrained optimization in N-P3O mitigates unsafe behaviors and improves both efficiency and stability, leading to the possibility of more reliable real-world deployment.
\begin{table}[htb!]
\begin{tabular}{ccccc}
\toprule
Method & Succ. & Max Ball Vel. & Max Ang. Vel. & Energy Cost \\
\midrule
N-P3O    &       \textbf{79.5\%}       & 1.05 m/s         &         \textbf{2.13 rad/s}         &           \textbf{108.6 J/s}         \\
PPO-1.0  &        64.8\%       &        \textbf{1.41 m/s}         &         3.21 rad/s         &       255.8 J/s             \\
PPO-1.5  &        43.4\%    &       1.21 m/s         &       2.86 rad/s         &       185.7 J/s             \\
\bottomrule
\end{tabular}
\caption{Comparison between N-P3O and PPO with fixed regularization coefficients of 1.0 and 1.5. The bold numbers indicate the best results under each metric.}
\label{tab:constrainedRL}
\vspace{-10pt}
\end{table}

\paragraph{Student adaptation}
Finally, we evaluate the role of the final student adaptation stage by comparing policies before and after online adaptation. As shown in Table~\ref{tab:adaption}, the adapted student achieves a success rate of 79.5\% with a kick accuracy of 0.956, while also reducing energy consumption to $110.78 \text{ J/s}$. In contrast, the pre-adaptation student attains only 52.3\% success and 0.807 accuracy, consuming more than twice as much energy ($256.2 \text{ J/s}$). For reference, the privileged teacher policy trained without sensor noise achieves slightly higher success (81.1\%). These results highlight the fact that online adaptation is essential for handling partially observed states that cannot be addressed by distillation of a privileged teacher policy.
\begin{table}[htb!]
\centering
\begin{tabular}{cccc}
\toprule
Method & Succ. & Kick Accuracy & Energy Cost \\
\midrule
Student after adaptation    &       79.5\%       &    \textbf{0.956}       &        \textbf{110.8 J/s}             \\
Student before adaptation    &     52.3\%       &     0.807           &     256.2 J/s                 \\
Teacher (w/o sensor noise) & \textbf{81.1\%}  & 0.954 & 240.1 J/s \\
\bottomrule
\end{tabular}
\caption{Evaluation of student policies before and after adaptation, and a teacher policy with perfect perception. The bold numbers indicate the best results under each metric.}
\label{tab:adaption}
\vspace{-20pt}
\end{table}